\documentclass[11pt,a4paper]{article}
\usepackage[hyperref]{acl2020}
\usepackage{times}
\usepackage{latexsym}
\usepackage{amsmath,amssymb,amsfonts}
\usepackage{comment}
\usepackage{hyperref}
\usepackage{verbatim}
\usepackage{enumitem}
\usepackage{subcaption}
\usepackage{cleveref}
\usepackage{footmisc}
\usepackage{lipsum}
\usepackage{graphicx}
\graphicspath{{imgs/}}

\newcommand\blfootnote[1]{%
  \begingroup
  \renewcommand\thefootnote{}\footnote{#1}%
  \addtocounter{footnote}{-1}%
  \endgroup
}

\usepackage{microtype}

\aclfinalcopy 
\def\aclpaperid{9} 

\title{Examination and Extension of Strategies for Improving Personalized Language Modeling via Interpolation}

\author{
{\bfseries Liqun Shao$^*$, Sahitya Mantravadi$^*$, Tom Manzini$^*$, Alejandro Buendia$^*$,} \\
{\bfseries Manon Knoertzer$^*$, Soundar Srinivasan, and Chris Quirk}\\
Microsoft Corp.\\
\textit {\{lishao,samantr,thmanzin,albuendi,maknoert,sosrini,chrisq\}@microsoft.com}
}
\date{}

\begin{document}
\maketitle
\begin{abstract}
 
In this paper, we detail novel strategies for interpolating personalized language models and methods to handle out-of-vocabulary (OOV) tokens to improve personalized language models.
Using publicly available data from Reddit, we demonstrate improvements in offline metrics at the user level by interpolating a global LSTM-based authoring model with a user-personalized \(n\)-gram model.
By optimizing this approach with a back-off to uniform OOV penalty and the interpolation coefficient, we observe that over 80\% of users receive a lift in perplexity, with an average of $5.2\%$ in perplexity lift per user.
In doing this research we extend previous work in building NLIs and improve the robustness of metrics for downstream tasks.

\end{abstract}

\section{Introduction}

\blfootnote{* Indicates equal contributions}


Natural language interfaces (NLIs) have become a ubiquitous part of modern life.
Such interfaces are used to converse with personal assistants (e.g., Apple Siri, Amazon Alexa, Google Assistant, Microsoft Cortana), to search for and gather information (Google, Bing), and to interact with others on social media.
One developing use case is to aid the user during composition by suggesting words, phrases, sentences, and even paragraphs that complete the user's thoughts~\cite{radford2019language}.

Personalization of these interfaces is a natural step forward in a world where the vocabulary, grammar, and language can differ hugely user to user \cite{ishikawa2015gender,rabinovich2018native}.
Numerous works have described personalization in NLIs in audio rendering devices \cite{morse2008system}, digital assistants \cite{chen2014personalized}, telephone interfaces \cite{partovi2005method}, etc. 
We explore an approach for personalization of language models (LMs) for use in downstream NLIs on composition assistance, 
and replicate previous work to show that interpolating a global long short-term memory network (LSTM) model with user-personalized \(n\)-gram models provides per-user performance improvements when compared with only a global LSTM model \citep{chen2015investigation,chen2019gmail}. 
We extend that work by providing new strategies to interpolate the predictions of these two models. 
We evaluate these strategies on a publicly available set of Reddit user comments and show that our interpolation strategies deliver a $5.2\%$ perplexity lift.
Finally, we describe methods for handling the crucial edge case of out-of-vocabulary (OOV) tokens\footnote{To the best knowledge of the authors, these edge cases are not clearly defined in the literature when combining two LMs trained on two different datasets.}.

Specifically, the contributions of this work are:
\begin{enumerate}
    \item We evaluate several approaches to handle OOV tokens, covering edge cases not discussed in the LM personalization literature.
    \item We provide novel analysis and selection of interpolation coefficients for combining global models with user-personalized models.
    \item We experimentally analyze trade-offs and evaluate our personalization mechanisms on public data, enabling replication by the research community.
\end{enumerate}

\section{Related Work} \label{sec:related work}

Language modeling is a critical component for many NLIs, and personalization is a natural direction to improve these interfaces.  

Several published works have explored personalization of language models using historical search queries \citep{jaech2018personalized}, features garnered from social graphs \cite{wen2012personalized,tseng2015personalizing, lee2016personalizing}, and transfer learning techniques \citep{yoon2017efficient}.
Other work has explored using profile information (location, name, etc.) as additional features to condition trained models \cite{shokouhi2013learning, jaech2018personalized}.
Specifically, in the NLI domain, Google Smart Compose \citep{chen2019gmail} productized the approach described in \citep{chen2015investigation} by using a linear interpolation of a general background model and a personalized \(n\)-gram model to personalize LM predictions in the email authoring setting. 
We view our work as a natural extension to this line of research because strategies that improve personalization at the language modeling level drive results at the user interface level.




\section{Personalized Interpolation Model}
\label{sec:interpolation}


The goal of text prediction is strongly aligned with language modeling.
The task of language modeling is to predict which words come next, given a set of context words.
In this paper, we explore using a combination of both large scale neural LMs and small scale personalized \(n\)-gram LMs. This combination has been studied in the literature \cite{chen2015investigation} and has been found to be performant. We describe mechanisms for extending this previous work in this section. 
Once trained, we compute the perplexity of these models not by exponentiation of the cross entropy, but rather by explicitly predicting the probability of test sequences. In practice this model is to be used to rerank sentence completion sequences. As a result, it is impossible to ignore the observation of OOV tokens.


\subsection{Personalized \(n\)-gram LMs}


Back-off \(n\)-gram LMs \citep{kneser1995improved} have been widely adopted given their simplicity, and efficient parameter estimation and discounting algorithms further improve robustness \citep{chen2015investigation}. Compared with DNN-based models, \(n\)-gram LMs are computationally cheap to train, lightweight to store and query, and fit well even on small data---crucial benefits for personalization.
%
Addressing the sharp distributions and sparse data issues in $n$-gram counts is critical.
We rely on Modified Kneser-Ney smoothing~\citep{james2000modified}, which is generally accepted as one of the most effective smoothing techniques.

\subsection{Global LSTM}
\label{sec_lstm}
For large scale language modeling, neural network methods can produce dramatic improvements in predictive performance~\citep{LimitsOfLM2016}.
Specifically, we use LSTM cells~\citep{hochreiter1997long}, known for their ability to capture long distance context without vanishing gradients. 
By computing the softmax function on the output scores of the LSTM we can extract the LSTM's per-token approximation as language model probabilities.

\subsection{Evaluation}
We use perplexity (PP) to evaluate the performance of our LMs. PP is  a measure of how well a probability model predicts a sample, i.e., how well an LM predicts the next word. This can be treated as a branching factor. Mathematically, PP is the exponentiation of the entropy of a probability distribution. Lower PP is indicative of a better LM. 
We define lift in perplexity (PP lift) as 
\begin{equation}\label{eq:liftp}
\text{PP lift} = \dfrac{PP_{\text{global}}-PP_{\text{interpolated}}}{PP_{\text{global}}},
\end{equation}
where $PP_{\text{interpolated}}$ is the perplexity of the interpolated model and $PP_{\text{global}}$ is the perplexity of the global LSTM model, which serves as the baseline. Higher PP lift is indicative of a better LM.

\subsection{Interpolation Strategies}
\label{interpolation_sec}
Past work \citep{chen2015investigation} has described mechanisms for interpolating global models with personalized models for each user. Our experimentation mixes a global LSTM model with the personalized \(n\)-gram models detailed above\footnote{We further detail the hyperparameters and training scheme of our LSTM and \(n\)-gram models in the appendices.}.

The interpolation is a linear combination of the predicted token probabilities:
\begin{equation}\label{eq:interpolation}
P = \alpha P_{\text{personal}} + (1-\alpha)P_{\text{global}}
\end{equation}

\noindent $\alpha$ indicates how much  personalization is added to the global model.
We explore constant values of $\alpha$, either globally or for each user. We compute a set of oracle $\alpha$ values, the values of $\alpha$ per user that empirically minimize interpolated perplexity. We compare our strategies for tuning $\alpha$ to these oracle $\alpha$ values, which present the best possible performance on the given user data in Section \ref{ssec:alpha-optimization}.
Intuitively, users whose comments have a high proportion of tokens outside the global vocabulary will need more input from the global model than their own personalized model to accurately model their language habits. Thus, we also explore an inverse relationship between $\alpha$ and each user's OOV rate. 

\subsection{OOV Mitigation Strategies}\label{subsec:oov}
When training on datasets with a large proportion of OOV tokens, low PP may not indicate a good model.
Specifically, if the proportion of OOV tokens in the data is high, the model may assign too much mass to OOV tokens resulting in a model with a propensity to predict the OOV token.
Such a model may have low PP, but only because it frequently predicts the commonly occurring OOV token.
While this may be an effective model of the pure sequence of tokens, it does not align with downstream objectives present at the interface level which relies on a robust prediction of non-OOV tokens.
Because of this disconnect between model and overall task objective, mitigation strategies must be implemented in order to adequately evaluate the performance of LMs in high OOV settings.
We evaluate the following strategies to mitigate this behavior:
\begin{enumerate}
    \item Do nothing, assigning OOV tokens their estimated probabilities;
    \item Skip the OOV tokens, scoring only those items known in the training vocabulary; and
    \item Back-off to a uniform OOV penalty, assigning a fixed probability $\phi$ to model the likelihood of selecting the OOV token\footnote{We consider $\phi$ to be a hyperparameter which must be tuned for each use case. In our experiments we assign $\phi$ to be $\frac{1}{V}$, where $V$ is the vocabulary size.}.
\end{enumerate}

When reporting our results we denote PP\textsubscript{base} as PP observed when using strategy 1, PP\textsubscript{skip} as PP observed when using strategy 2, and PP\textsubscript{backoff} as PP observed when using strategy 3. 

\section{Data}
\begin{figure}[h]
  \includegraphics[width=\columnwidth,height=\textheight,keepaspectratio]{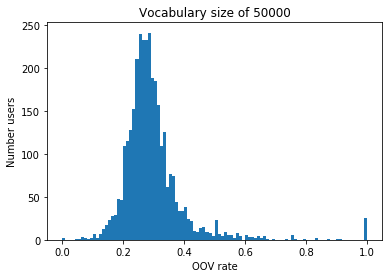}
  \caption{Histogram of OOV rates for 3265 users' training data with a vocabulary size of 50,000.}
  \label{fig:oov_vocab_50k}
\end{figure}

\begin{figure}[h]
  \includegraphics[width=\columnwidth,height=\textheight,keepaspectratio]{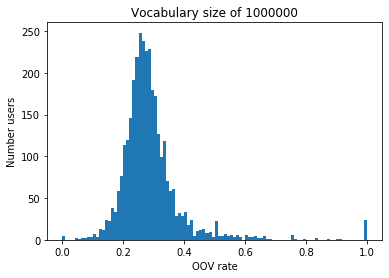}
  \caption{Histogram of OOV rates for 3265 users' training data with a vocabulary size of 1,000,000.}
  \label{fig:oov_vocab_1m}
\end{figure}

The data for our model comes from comments made by users on the Internet social media website Reddit\footnote{We retrieved copies of \href{www.reddit.com}{www.reddit.com} user comments from \href{https://pushshift.io/}{https://pushshift.io/}.}. Reddit is a  rich source of natural-language data with high linguistic diversity due to posts about a variety of topics, informality of language, and sheer volume of data.
As a linguistic resource, Reddit comments present in a heavily conversational and colloquial tone, and users frequently use slang and misspell words. Because of this there are a high number of unique tokens. As developers of a machine learning system, we seek to balance having a large vocabulary in order to capture the most data with having a small vocabulary in order to keep the model from overfitting. We construct our vocabulary by empirically selecting the $n$ most common tokens observed by randomly selecting Reddit user comments. We then share this vocabulary, created from the global training set, in both the personalized and global models. This value of $n$ must be tuned based on data. When choosing a size for vocabulary, there exists a tradeoff between performance and capturing varied language. Larger vocabularies adversely impact performance but may encapsulate more variability of language. For a given vocabulary size chosen from training data for the global LSTM, we plot the resulting OOV rates for users. As can be seen when comparing Figure \ref{fig:oov_vocab_50k} and Figure \ref{fig:oov_vocab_1m}, very few gains in user-level OOV rates are seen when expanding the vocabulary size twenty-fold. Thus, we choose a vocabulary size of 50,000.

\label{sec:supplemental}



For the global LSTM, we split the global distribution of Reddit data into training sourced from 2016, validation sourced from 2017, and test sourced from 2018.
We sampled such that \(70\%\) of users were reserved for training, \(20\%\) of users for validation, and \(10\%\) of users for test. We allot \(100,000\) users for the test set and scale the number of users in the other sets accordingly.
There are 10 billion total tokens in the training data, with 29 million unique tokens. 90\% of unique tokens occur 6 or fewer times, and half of users have 20 or fewer comments per year with an average comment length of 13 tokens.
For the personalized \(n\)-grams, we selected all comment data from 3265 random Reddit users\footref{supplemental} who made at least one comment in each of 2016, 2017, and 2018. Then, for each user, we selected the data from 2016 as training data, the data from 2017 as validation data, and the data from 2018 as testing data. 

\section{Results}
\label{sec:results}
Here we discuss the results observed when evaluating the interpolated global LSTM and user-personalized \(n\)-gram model on users' comments using various OOV mitigation and $\alpha$ interpolation strategies.

\subsection{OOV Mitigation Strategies}  
In our data used for personalization, 68\% users have more than 25\% OOV rate for validation data, and 65\% users have more than 25\% OOV rate for training data. This  empirically causes large deviations between the different PP\textsubscript{backoff}, PP\textsubscript{skip}, and PP\textsubscript{base}.
We find that a personalized \(n\)-gram model can't handle OOV tokens very well in high OOV settings, because it assigns higher probabilities to OOV tokens than some of the tokens in the vocabulary.
As discussed in Section \ref{subsec:oov} high OOV rates at the per-user level PP\textsubscript{base} present a view of the results that is disconnected from downstream use in an NLI.
At the same time, PP\textsubscript{skip} presents the view most aligned with the downstream task because in an NLI the OOV token should never be shown.
However, PP\textsubscript{skip} comes with some mathematical baggage.
Specifically, when all tokens are OOV, the PP\textsubscript{skip} will be infinite.
These two approaches represent the extremes of the strategies which could be used.
We argue that PP\textsubscript{backoff} represents the best of both worlds.

\begin{figure}[h]
  \includegraphics[width=\columnwidth,height=\textheight,keepaspectratio]{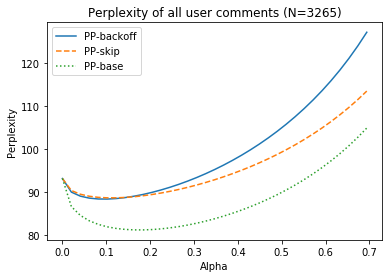}
  \caption{Average of interpolated PP for all users for varied values of $\alpha$ $\leq$ 0.7 for each method of approaching OOV tokens.}
  \label{fig:interpolated_pps}
\end{figure}

Figure \ref{fig:interpolated_pps} shows that PP\textsubscript{backoff} provides measurements near the minima that are closely aligned with PP\textsubscript{skip} while also being free of the mathematical and procedural issues associated with PP\textsubscript{skip} and PP\textsubscript{base}.
We provide an example to further illustrate the above statement. Consider a high OOV rate comment such as ``re-titled jaff ransomware only fivnin.” with OOV tokens \textit{re-titled, jaff, ransomware, fivnin}. Following encoding, the mode would see this sequence as ``OOV OOV OOV only OOV". When measuring the probability of this sequence a model evaluated using PP\textsubscript{base} would have lower perplexity because it has been trained to overweight the probability of OOV tokens as they occur more frequently than the tokens they represent. However, this sequence should have far lower probability, and thus higher perplexity, because the model is in fact failing to adequately model the true sequence.
We argue that assigning a uniform value $\theta$ to OOV tokens will more accurately represent the performance of the model when presented with data with a high quantity of OOV tokens.

Because we believe that PP\textsubscript{backoff} presents the most accurate picture of model performance, we have chosen to present our results in Section \ref{ssec:user_analysis_results} and \ref{ssec:alpha-optimization} using PP\textsubscript{backoff}.

\subsection{Analysis of Personalization}
\label{ssec:user_analysis_results}
\begin{figure}[h]
  \includegraphics[width=\columnwidth,height=\textheight,keepaspectratio]{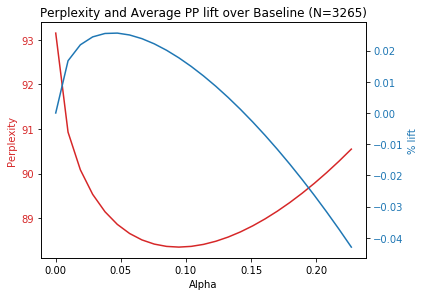}
  \caption{PP\textsubscript{backoff} and average PP\textsubscript{backoff} lift over baseline for various values of $\alpha < 0.22$.}
  \label{fig:alpha_pp_lift}
\end{figure}

We next present an interesting dichotomy in Figure \ref{fig:alpha_pp_lift} not previously discussed in the personalization literature. 
In the constant $\alpha$ for all users setting we can optimize to either minimize the overall PP\textsubscript{backoff} for all users or to maximize the average PP\textsubscript{backoff} lift across users. 
These two objectives result in different constant $\alpha$\footnote{There may be other trade-offs to examine.}. 
Specifically, minimizing PP\textsubscript{backoff} over users yields $\alpha = 0.105$, providing an improvement for $67.3\%$ of users and an average PP\textsubscript{backoff} lift of $2.5\%$.
Maximizing the average PP\textsubscript{backoff} lift per user yields $\alpha = 0.041$, providing an improvement for $74.2\%$ of users and an average PP\textsubscript{backoff} lift of $2.7\%$\footnote{Further details are included in the appendices.\label{supplemental}}.

\subsection{Constant and Personalized Interpolation Coefficient $\alpha$ Optimization}
\label{ssec:alpha-optimization}
When searching for a constant value for $\alpha$ for all users, $\alpha = 0.105$ achieves the minimum mean interpolated PP\textsubscript{backoff}, with an average PP\textsubscript{backoff} lift of $2.5 \%$.

Next, we personalize the value of $\alpha$ for each user. We first produce a set of oracles\footref{supplemental} as described in Section \ref{interpolation_sec}. With this set of oracle values of $\alpha$, the average PP\textsubscript{backoff} lift is $6.1 \%$ with the best average PP\textsubscript{backoff} achievable in this context. While it is possible to compute the oracle values for each user in a production setting, this may not be tractable when user counts are high and there exist latency constraints.

\begin{figure}[h]
  \includegraphics[width=\columnwidth,height=\textheight,keepaspectratio]{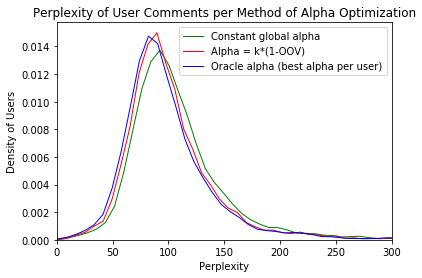}
  \caption{Distribution of interpolated PP\textsubscript{backoff} for users using each method of $\alpha$ optimization. The values for $\alpha = k\cdot (1 - \text{OOV rate})$ are averaged over 10 random selections.}
  \label{fig:alpha_pp_distr}
\end{figure}

Thus, we try an inverse linear relationship: $\alpha = k\cdot (1 - \text{OOV rate})$. To illustrate the effect of this relationship, we perform this optimization 10 times, using a different random subset of users each time to optimize $k$ and then evaluate on the rest of the users. On average, we observe a PP\textsubscript{backoff} lift of $5.2 \%$, and $80.1 \%$ of users achieve an improvement in PP\textsubscript{backoff}.
In Figure \ref{fig:alpha_pp_distr} we see that a heuristic approach of lower complexity achieves near-oracle performance, with the distribution of PP\textsubscript{backoff} for this method closely matching the oracle distribution of PP\textsubscript{backoff}. 
We also find that this method of $\alpha$ personalization yields lower PP\textsubscript{backoff} for more users than using a constant value for $\alpha$. 

\section{Conclusion \& Future Work}
\label{sec:conclusion}
In this paper we presented new strategies for interpolating personalized LMs, discussed strategies for handling OOV tokens to give better vision into model performance, and evaluated these strategies on public data allowing the research community to build upon these results.
Furthermore, two directions could be worth exploring: Investigate on when personalization is useful at a user level to better interpret the results; Research on user-specific vocabularies for personalized models instead of using a shared vocabulary for both the personalized and global background models.

As NLIs move closer to the user, personalization mechanisms will need to become more robust.
We believe the results we have presented form a natural step in building that robustness.


\section*{Acknowledgments}
We would like to thank the Microsoft Search, Assistant and Intelligence team, and in particular Geisler Antony, Kalyan Ayloo, Mikhail Kulikov, Vipul Agarwal, Anton Amirov, Nick Farn and Kunho Kim, for their invaluable help and support in this research. We also thank T. J. Hazen, Vijay Ramani, and the anonymous reviewers for their insightful feedback and suggestions.

\newpage
\bibliography{anthology,acl2020}
\bibliographystyle{acl_natbib}

\appendix
\section{Appendices}
\label{sec:appendix}
\subsection{Hyperparameters and Model Training}
The global LSTM model trained token embeddings of size 300, and had hidden unit layers of size 256 and 128, an output projection of dimension 100, and a vocabulary of 50,000 tokens. It was trained with dropout using the Adam optimizer, and we parallel-trained our global LSTM on an Azure\footnote{\href{www.azure.com}{www.azure.com}} \text{Standard\_NC24s\_v2} machine which includes \(24\) vCPUs and \(4\) NVIDIA Tesla P100 GPUs.

The personalized \(n\)-gram models were \(3\)-gram modified Kneser-Ney smoothed models with discounting values of 0.5 (\(1\)-grams), 1 (\(2\)-grams), and 1.5 (\(3\)-grams).

\subsection{User Analysis Plots}
\label{subsec:user_analysis}
The average size of the user-personalized corpus is around 140 comments, while the median size is 23 comments. The average comment length for each user is around 14 tokens.

By analyzing the results with the lowest interpolated PP\textsubscript{backoff} ($\alpha=0.105$ for all users), we make two observations: users with average comment length less than around 30 tokens don’t get much benefit from personalization, and users with less than around 100 comments don’t get much benefit from personalization. 
\begin{figure}[]
  \includegraphics[width=\columnwidth,height=\textheight,keepaspectratio]{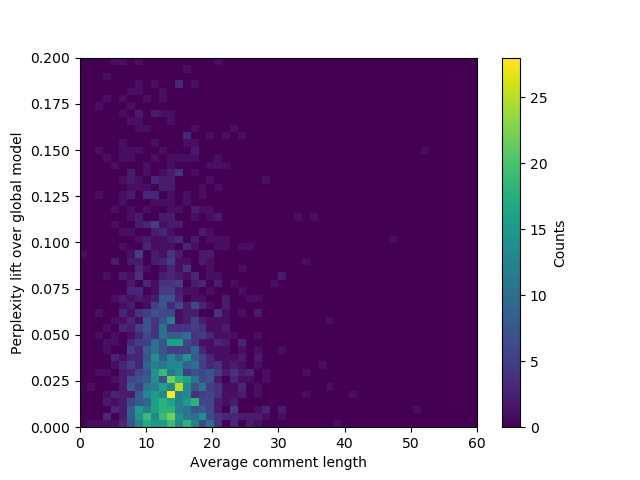}
  \caption{Histogram of PP lift over global model vs. average comment length ($\alpha=0.105$).}
  \label{fig:aver_comm}
\end{figure}
\begin{figure}[]
  \includegraphics[width=\columnwidth,height=\textheight,keepaspectratio]{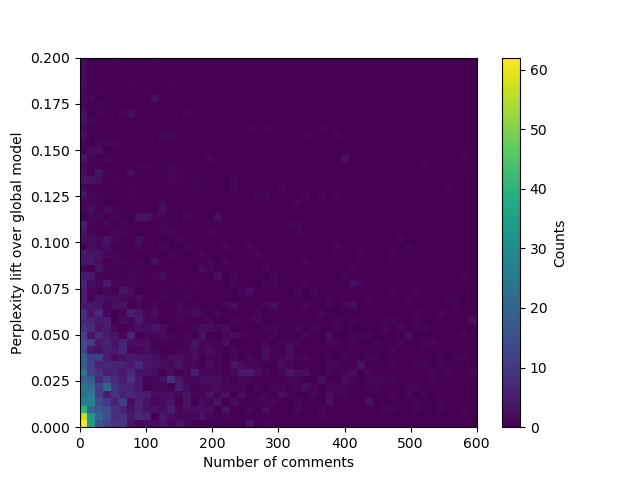}
  \caption{Histogram of PP lift over global model vs. number of comments ($\alpha=0.105$).}
  \label{fig:num_comm}
\end{figure}
\vfill\eject
\subsection{Oracle $\alpha$ Distribution}
\label{subsec:oracle_alphas}
\begin{figure}[h]
  \includegraphics[width=\columnwidth,height=\textheight,keepaspectratio]{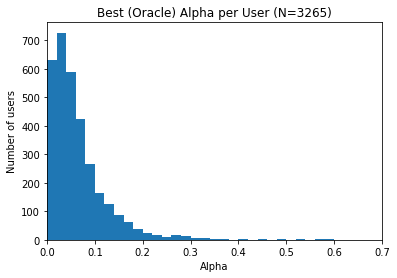}
  \caption{Distribution of oracle values of $\alpha$ per user.}
  \label{fig:oracle_alphas}
\end{figure}

Figure \ref{fig:oracle_alphas} shows the distribution of the empirically-computed ``oracle" values for $\alpha$.
  

\end{document}